\documentclass[a4paper,conference]{IEEEtran}
\IEEEoverridecommandlockouts
\usepackage{cite}
\usepackage{amsmath,amssymb,amsfonts}
\usepackage{algorithmic}
\usepackage{graphicx}
\usepackage{textcomp}
\usepackage{xcolor}

\usepackage{graphics}
\usepackage{multirow}
\usepackage{hyperref}
\usepackage[utf8]{inputenc}
\DeclareUnicodeCharacter{FB01}{fi}
\usepackage[small]{caption}
\usepackage{booktabs}
\usepackage{algorithm}
\usepackage{algorithmic}
\usepackage{amsfonts,amssymb}
\def\BibTeX{{\rm B\kern-.05em{\sc i\kern-.025em b}\kern-.08em
    T\kern-.1667em\lower.7ex\hbox{E}\kern-.125emX}}
\begin{document}
\title{PIN: A Novel Parallel Interactive Network for Spoken Language Understanding}

\author{\IEEEauthorblockN{1\textsuperscript{st} Peilin Zhou}
\IEEEauthorblockA{\textit{ADSPLAB} \\
\textit{Peking University}\\
Shenzhen, China \\
zhoupl@pku.edu.cn}
\and
\IEEEauthorblockN{2\textsuperscript{nd} Zhiqi Huang}
\IEEEauthorblockA{\textit{ADSPLAB} \\
\textit{Peking University}\\
Shenzhen, China \\
zhiqihuang@pku.edu.cn}
\and
\IEEEauthorblockN{3\textsuperscript{rd} Fenglin Liu}
\IEEEauthorblockA{\textit{ADSPLAB} \\
\textit{Peking University}\\
Shenzhen, China \\
fenglinliu98@pku.edu.cn}
\and
\IEEEauthorblockN{4\textsuperscript{th} Yuexian Zou\thanks{\IEEEauthorrefmark{1} is corresponding author.}\IEEEauthorrefmark{1}}
\IEEEauthorblockA{\textit{ADSPLAB} \\
\textit{Peking University}\\
Shenzhen, China \\
zouyx@pku.edu.cn}}


%Telephone: (800) 555--1212, Fax: (888) 555--1212}


\maketitle

\begin{abstract}
% 邹老师修改版
Spoken Language Understanding (SLU) is an essential part of the spoken dialogue system, which typically consists of intent detection (ID) and slot filling (SF) tasks. Recently, recurrent neural networks (RNNs) based methods achieved the state-of-the-art for SLU. It is noted that, in the existing RNN-based approaches, ID and SF tasks are often jointly modeled to utilize the correlation information between them. However, we noted that, so far, the efforts to obtain better performance by supporting bidirectional and explicit information exchange between ID and SF are not well studied.
% However, we note that, so far, the explicit and bidirectional information flow for ID and SF tasks has not been explored to improve the performance of SLU. 
% In addition, the utilization of the local context information will enhance the performance of SF. 
In addition, few studies attempt to capture the local context information to enhance the performance of SF. Motivated by these findings, in this paper, Parallel Interactive Network (PIN) is proposed to model the mutual guidance between ID and SF. Specifically, given an utterance, a Gaussian self-attentive encoder is introduced to generate the context-aware feature embedding of the utterance which is able to capture local context information. Taking the feature embedding of the utterance, Slot2Intent module and Intent2Slot module are developed to capture the bidirectional information flow for ID and SF tasks. Finally, a cooperation mechanism is constructed to fuse the information obtained from Slot2Intent and Intent2Slot modules to further reduce the prediction bias.
The experiments on two benchmark datasets, i.e., SNIPS and ATIS, demonstrate the effectiveness of our approach, which achieves a competitive result with state-of-the-art models. More encouragingly, by using the feature embedding of the utterance generated by the pre-trained language model BERT, our method achieves the state-of-the-art among all comparison approaches.
\end{abstract}


% \fi
\IEEEpeerreviewmaketitle

\section{Introduction}
\begin{figure}[t]
\centerline{\includegraphics[width=9 cm]{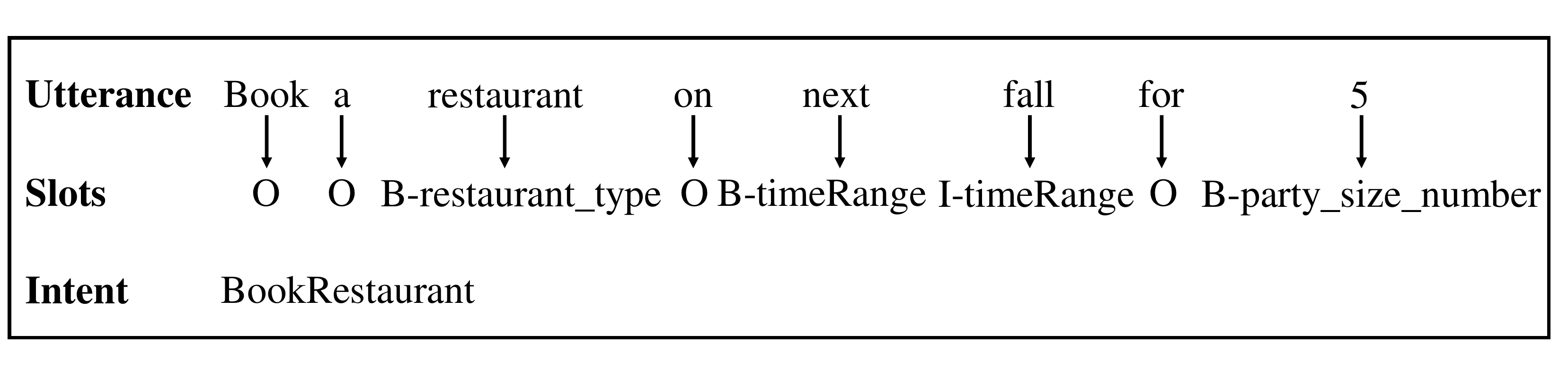}}
\caption{An example of an utterance with BIO format annotations for slot filling, which indicates the slot of restaurant type, time range, and party size number from an utterance with an intent BookRestaurant.}
\label{fig:example}
\end{figure}

Spoken Language Understanding (SLU) technology plays a crucial part in goal-oriented dialogue systems. It typically involves intent detection (ID) and slot filling (SF) tasks.
As the names imply, intent detection aims to identify users’ intents, while slot filling focuses on capturing semantic constituents from user utterances  \cite{tur2011spoken}. As shown in Fig.~\ref{fig:example}, given a user query \textit{‘Book a restaurant on next fall for 5’}, which is sampled from the SNIPS dataset \cite{coucke2018snips}, intent BookRestaurant is assigned to the whole sentence, and each token in the sentence corresponds to one specific slot type. Due to the process interdependence between SLU and subsequent dialogue components, such as the dialogue manager and the natural language generator, performance on these two tasks, i.e., ID and SF, determines the upper limit for the utility of such dialogue system \cite{chen2017survey}.

\begin{figure}[t]
\centerline{\includegraphics[width=9 cm]{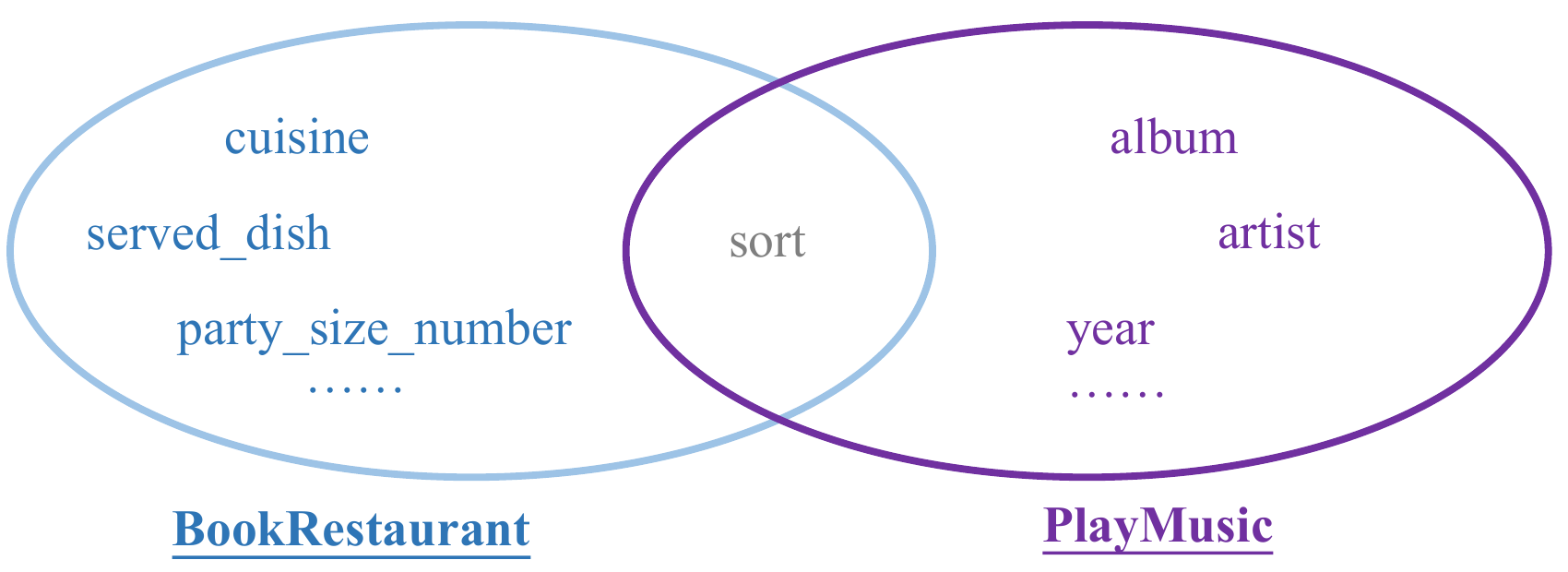}}
\caption{An example of the co-occurrence characteristic between slot tags and intent labels. The underlined text represents the intent label, and texts inside the circle represent the slot tags correspond to that intent.}
\label{fig:cooccurence}
\end{figure}

Intuitively, intent detection and slot filling are associated with each other  \cite{goo2018slot}, which can be observed in Fig.~\ref{fig:cooccurence}.
For instance, when the intent of an utterance is \textit{PlayMusic}, the slots of the utterance are more likely to be \textit{artist} rather than \textit{cuisine}, and vice versa 
% (i.e., the more topic-specific slot types appear in the utterance, the greater the likelihood of a certain intent assigned). 
(i.e., The more topic-related slots appear in an utterance, the more likely the intent of the utterance corresponds to that topic.).
As the accumulation of annotated queries, the co-occurrence characteristic between slot tags and intent labels can become more prominent and perceptible, providing hints about the mutual dependence of ID and SF.
Hence, it is promising to achieve a complementary effect by modeling the two tasks in a joint fashion and sharing knowledge between them.
% \cite{xu2013convolutional} proposed using CNN based triangular CRF for joint intent detection and slot filling. 
% Some works \cite{liu2016attention,zhang2016joint,hakkani2016multi} simply rely on the shared parameters to model this co-occurrence characteristic in an implicit way. 
Some works \cite{liu2016attention,zhang2016joint,hakkani2016multi} proposed to model intent-slot relation by sharing parameters, outperforming previous separated models by a large margin.
% With the rise of RNN-based methods and attention mechanisms, the practice of working the relationship between intents and slots into joint models is likely to get more sophisticated.
More recently, gate mechanism and attention mechanism were also introduced to the RNN-based models \cite{goo2018slot} \cite{li2018self} , which provides a new perspective for joint ID and SF modeling. 
% \cite{goo2018slot} proposed using a slot-gated mechanism to enhance slot filling performance with intent information. To take one step further, \cite{qin2019stack} proposed a Stack-Propagation Framework to incorporate token-level intent information to better guide the slot prediction process. This stacking neural network model could provide better interpretability than the slot-gated mechanism.

However, these methods still suffer from various limitations. 
For one thing, local context information is not fully exploited in their models, ignoring the intuition that local context is a useful architectural inductive prior for SF. For another thing, most methods fail to take full advantage of the supervised signals due to their implicit or unidirectional modeling style of the intent-slot relations. 
Those limitations will hinder the further improvement of SLU systems, especially the overall accuracy, which highly depends on the joint performance of ID and SF.  

In this work, we propose a novel Parallel Interactive Network (PIN) to address above issues. For the first issue, a Gaussian self-attentive encoder~\cite{guo2019gaussian} is introduced to better capture local structure and contextual information at each token, which incorporates valuable inductive prior knowledge for SF. For the second issue, we design a Intent2Slot module and a Slot2Intent module  to model the bidirectional information flow between SF and ID. Specifically, inspired by the Dual Process Theory (DPT)~\cite{dualprocess} in neurocognitive science, we divide the information processing in these modules into two stages: the implicit interaction stage and the explicit interaction stage. These two stages correspond to two different processing styles in which the human brain operates: implicit (intuitive), unconscious learning and explicit (rational), conscious learning. In the implicit interaction stage, the relationships between intents and slots are implicitly captured in the parameters of the shared encoder, which is then utilized by the intuitive decoders to obtain token-level intent distribution and slot label distribution. In the explicit interaction stage, those distribution information obtained in former stage is explicitly utilized by rational decoders to reduce the solution space.
Finally, a cooperation mechanism, which comprehensively considers information from above two stages, is performed to reduce the prediction bias and thereby improve the precision and accuracy of model predictions.

To verify the effectiveness of our proposed method, we conduct experiments on two real-world datasets, i.e., ATIS \cite{hemphill1990atis} and SNIPS \cite{coucke2018snips}, which are popularly used as benchmarks in recent works. Empirical results show that our method achieves competent performance on intent error rate, slot F1-score, and sentence-level semantic frame accuracy compared with other baselines.
% 我门还使用bert作为预训练模型，进一步提升了模型的表现。
In addition, Bidirectional Encoder Representation from Transformer (BERT) \cite{devlin2018bert} is explored to further improve the performance of our model.

In summary, the key contributions are as follows:
\begin{itemize}
\item We propose a novel parallel interactive network (PIN), which divides the mutual guidance between ID and SF into two interaction stages, i.e., implicit interaction stage and explicit interaction stage, to improve the performance and interpretability of our approach.
\item We propose a novel cooperation mechanism within the PIN model in order to effectively combine and balance the information provided by the two interaction stages. It can further refine the prediction results of the proposed model and alleviate the error propagation problem.
\item We validate our approach on two benchmark datasets. The experimental results  demonstrate the effectiveness of our approach, which outperforms all comparison methods in terms of most metrics on the two publicly benchmark datasets. 
\end{itemize}
\section{Related work}

\subsection{Intent Detection}

In recent years, most researchers treat intent detection as a Semantic Utterance Classification (SUC) problem \cite{dauphin2013zero}. The early traditional method is to use rule-based templates \cite{dowding1994gemini} to match the most appropriate intent, which is extremely inflexible and effort-consuming if intent categories changed. 
Some statistical-based methods,  such as Naive Bayes  \cite{mccallum1998comparison}, Adaboost \cite{schapire2000boostexter}, Support Vector Machine (SVM) \cite{haffner2003optimizing} and logistic regression \cite{genkin2007large}, have also been explored to improve the performance of intent detection. But these approaches still have difficulty in understanding the deep semantics of user utterances due to the ambiguity and irregularity of user expressions. With the rise of deep learning techniques, many neural network based models are proposed to better solve this classification problem. \cite{deoras2013deep} attempted to use Deep belief networks (DBNs) in call routing classification. \cite{kim2014convolutional,zhang2015sensitivity,hashemi2016query} proposed using CNN to extract features of sentences and has achieved excellent results. 
More recently, \cite{ravuri2015recurrent} adopted GRU and LSTM to capture the long-range dependency between words, which showed excellent performance on the intent detection problem. Besides, \cite{xia2018zero} demonstrated that capsule network could also be applied to this task.

\subsection{Slot Filling}

Typically, slot filling is often regarded as a sequence labelling task. Compared with sentence-level intent detection, word-level slot filling heavily relies on fine-grained semantic features. Traditional rule-based approaches directly extract semantic concepts from the user utterance based on hand-crafted rules \cite{kim2000rule}. They are in the throes of poor generalization and bad transferability.
Another line of works centered on feature-based supervised learning algorithms \cite{li2018survey}, such as Maximum Entropy Models \cite{kapur1989maximum}, Hidden Markov Models (HMM) \cite{eddy1996hidden}, and Conditional Random Fields (CRF) \cite{lafferty2001conditional}. Though achieving better performance than rule-based models, they still suffered from designing elaborate features, which is obviously time-consuming and tedious.
Recently, DL-based models for SF have taken a leading role and achieved state-of-the-art results. Bi-directional and hybrid RNN \cite{mesnil2013investigation}, deep LSTM \cite{yao2014spoken}, RNN-EM \cite{peng2015recurrent}, attention-based encoder-decoder \cite{simonnet2015exploring}, CNN \cite{vu2016sequential}, and joint pointer and attention \cite{zhao2018improving} are some typical works of this research direction.
\begin{figure*}[htbp]
\centerline{\includegraphics[width=18 cm]{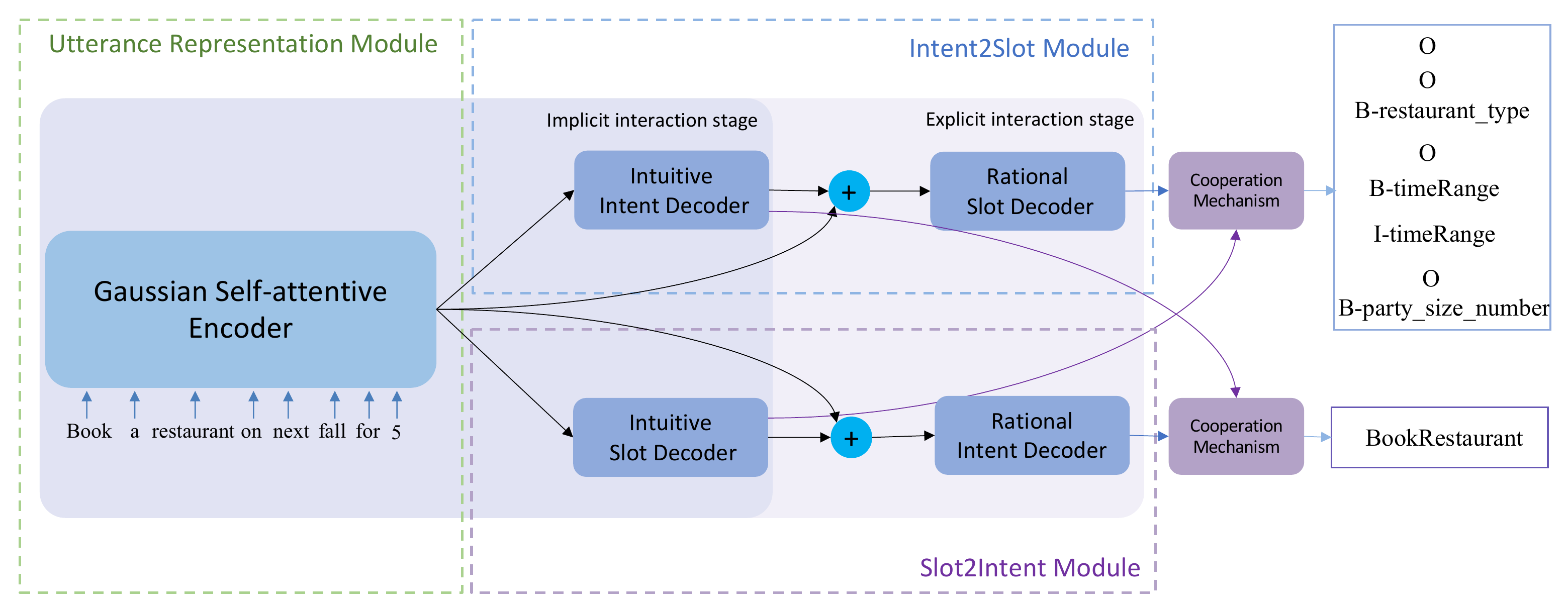}}
\caption{Illustration of our Parallel Interactive Network (PIN) for joint intent detection and slot filling. The PIN consists of the Utterance Representation Module, the Intent2Slot Module and the Slot2Intent Module. For a given utterance, the Utterance Representation module will first read and encode it as context-aware text representation, which is then fed to the Intent2Slot Module and Slot2Intent Module to model the interaction between ID and SF in both implicit and explicit manners. Finally, a cooperation mechanism is constructed to fuse the information obtained from  Slot2Intent and Intent2Slot modules  to further reduce the prediction  bias.}

\label{fig:network}
\end{figure*}
\subsection{Joint Modelling}

Intuitively, intents and slots represent the semantics of user actions from different granularity perspectives. Hence, there have been some works attempting to jointly model intent detection and slot filling.  Hence, how to better model the interactions between slots and intents is the crux of the matter.
\cite{liu2016attention} proposed using an attention-based neural network to jointly model the two tasks. Without directly exchanging information between intents and slots, it simply relied on a joint loss function to ``implicitly'' consider both cues. 
\cite{goo2018slot} introduced a slot-gated mechanism to improve slot filling by conditioning on the learned intent results. \cite{li2018self} proposed using intent-augmented embedding as a special gate function to guide the process of slot filling. \cite{qin2019stack} proposed a Stack-Propagation framework to explicitly incorporate token-level intent information to slot filling, which can further reduce the error propagation. However, these unidirectional interaction approaches ignored the fact that slot filling results are also instructive to the intent detection task. Thus it is necessary to consider the cross-feeding interactions between intents and slots for these two tasks. For instance, \cite{haihong2019novel} proposed a bi-directional interrelated model to achieve the effects of mutual enhancement. \cite{zhang2018joint} adopted a hierarchical capsule network to leverage the hierarchical relationships among words, slots, and intents in an utterance. 
In contrast to these lines of work, we propose to explicitly model the mutual support in a parallel manner, which is more straightforward and does better in capturing the complicated interdependence among slots and intents. Moreover, empirical results on two benchmark datasets demonstrate the effectiveness of our method, which can further improve the performance of SLU systems.
% \section{Problem Formulation}
% In this section, we describe the formulation deﬁnition for intent detection and slot filling in SLU.
% \begin{itemize}
% \item \textbf{Intent Detection} Given an input utterance $\mathbf{X}=\left( {{{x}}_{1}},{{{x}}_{2}},\ldots ,{{{x}}_{T}} \right)$ consisting of a sequence of T tokens, intent detection can be seen as a utterance-level sequence classiﬁcation problem to decide the corresponding intent label ${o_{t}^{I}}$ for each utterance.
% \item \textbf{Slot Filling} Slot filling can be seen as a sequnece labelling task, which maps an input utterance $\mathbf{X}=\left( {{{x}}_{1}},{{{x}}_{2}},\ldots ,{{{x}}_{T}} \right)$ to slots sequence 
% \end{itemize}

\section{Approach}
% Fig.~\ref{fig:network} shows the overall architecture of our proposed model. For a given utterance, the Utterance Representation module will first read and encode it as context-aware text representation, which is then fed to the Intent2Slot Module and Slot2Intent Module to model the interaction between ID and SF in both implicit and explicit manners. Finally a cooperation mechanism is proposed to ensemble the results and further reduce the prediction bias.
\subsection{Utterance Representation Module}

% In our xxx, intent detection task and slot filling task share one encoder, In the self-attentive encoder, we use BiLSTM with self-attention mechanism to leverage both advantages of temporal features and contextual information, which are useful for sequence labeling tasks \cite{zhong2018global}.
In the utterance representation module, we use BiLSTM with Gaussian self-attention mechanism to leverage both advantages of local structure and contextual information for a given utterance, which are useful for ID and SF tasks.
%sequence labeling tasks.
%\cite{zhong2018global}.
% For a given utterance, the utterance representation module will first read and encode it as  context-aware  text  representation. In practice, we 

The BiLSTM~\cite{hochreiter1997long}, which consists of a forward LSTM and a backward LSTM, transforms each input utterance $\mathbf{X}=\left( {{{x}}_{1}},{{{x}}_{2}},\ldots ,{{{x}}_{T}} \right)$ into  hidden states $\mathbf{H}=\left( {{{h}}_{1}},{{{h}}_{2}},\ldots ,{{{h}}_{T}} \right)$ as follows: 
\begin{equation}
{{\vec{h}}_{t}}=\overrightarrow{LSTM}\left( {{\phi }^{\text{emb}}}\left( {{x}_{i}} \right),{{{\vec{h}}}_{t-1}} \right)
\end{equation}
\begin{equation}
{{\overset{\scriptscriptstyle\leftarrow}{h}}_{t}}=\overleftarrow{LSTM}\left( {{\phi }^{\text{emb}}}\left( {{x}_{i}} \right),{{{\overset{\scriptscriptstyle\leftarrow}{h}}}_{t+1}} \right)
\end{equation}
\begin{equation}
{{h}_{t}}={{\vec{h}}_{t}}\oplus {{\overset{\scriptscriptstyle\leftarrow}{h}}_{t}}
\end{equation}
where ${{\phi }^{\text{emb}}}\left( {{x}_{i}} \right)$ is the embedding of word ${{x}_{i}}$ and  ${\oplus }$ is concatenation operator.

Recently, self-attention mechanisms have found broad application in various NLP tasks such as machine translation \cite{vaswani2017attention}, language understanding \cite{devlin2018bert,radford2018improving}, and extractive summarization \cite{liu2019fine}. However, the vanilla self-attention mechanism used in Transformer~\cite{vaswani2017attention} treats the same words almost equally without considering their positions, which contradict with our intuition that adjacent words contribute more semantically to central words \cite{guo2019gaussian}. As a result, the vanilla self-attention mechanism fails to model the local structure of texts,
which can provide useful prior knowledge to the SF task.
%which is useful inductive prior for SF task. 
To mitigate this problem, we apply a Gaussian prior to self-attention mechanisms to better capture both local and contextual information for each token. Formally, we follow the recently proposed Gaussian Transformer~\cite{guo2019gaussian} to define the Gaussian self-attention function as:
\begin{equation}
{{c}_{i}}=\sum\limits_{j}{\operatorname{Softmax}}\left( -\left| wd_{i,j}^{2}+b \right|+\left( {{x}_{i}}\cdot {{x}_{j}} \right) \right){{x}_{j}}
\end{equation}
\begin{equation}
{\mathbf{C}=\left( {{c}_{1}},{{c}_{2}},\ldots ,{{c}_{T}} \right)}
\end{equation}
where ${{c}_{i}}$ is the contextual representation (i.e., context vector) for ${{i}}$-th token, ${{x}_{i}}$ and ${{x}_{j}}$ represents the ${{i}}$-th token (i.e., central token) and ${{j}}$-th token from the sentence $\mathbf{X}$, $d_{i,j}^{{}}$ represents the distance between tokens, $\left| \text{ }\cdot \text{ } \right|$ represents the absolute value, $w>0,b\le 0$ are scalar parameters. For more details about Gaussian attention, please refer to \cite{guo2019gaussian}.

The final utterance representation $\mathbf{E}$ is the concatenation of the output of Gaussian self-attention and BiLSTM:
\begin{equation}
{\mathbf{E}=\mathbf{H}\oplus \mathbf{C}}
\end{equation}
where $\mathbf{E}=\left( {{\mathbf{e}}_{1}},\ldots ,{{\mathbf{e}}_{T}} \right)\in {{\mathbb{R}}^{T\times 2d}}$ and ${{\mathbf{e}}_{i}}$ can grasp both local and contextual information for ${i}$ th token by Gaussian self-attention mechanism. $\mathbf{E}$ is then fed to subsequent Slot2Intent Module and Intent2Slot Module simultaneously.

\subsection{Slot2Intent Module} % 挺好的..

In our work, the Slot2Intent module is designed to explicitly incorporate slot signals to intent prediction process. Specifically, a intuitive slot decoder is used to decode the utterance representation E into slot label sequence. Then, these token-level slot features are fed to a rational intent decoder to guide the intent prediction process.
\begin{itemize}
\item \textbf{Intuitive Slot Decoder.} We use a unidirectional LSTM as the intuitive slot decoder. The decoder’s hidden vector at each decoding time step $t$ is calculated as:
\begin{equation}
{\mathbf{h}_{t}^{IS}=LSTM\left( \mathbf{h}_{t-1}^{IS},\mathbf{y}_{t-1}^{IS}\oplus {{\mathbf{e}}_{t}} \right)}
\end{equation}
\begin{equation}
{\mathbf{y}_{t}^{IS}=\operatorname{softmax}\left( \mathbf{W}_{h}^{IS}\mathbf{h}_{t}^{IS} \right)}
\end{equation}
where $\mathbf{h}_{t-1}^{IS}$ is the previous decoder state, ${{\mathbf{e}}_{t}}$ is the aligned encoder hidden state, $\mathbf{W}_{h}^{IS}$ are trainable parameters of the model, $\mathbf{y}_{t}^{IS}$ is the slot label distribution of the ${{t}}$-th token in the utterance.

% Then the hidden vector $\mathbf{h}_{t}^{IS}$ is utilized for slot filling: 
\item \textbf{Rational Intent Decoder.} To consider both the utterance representation and slot information, we use another unidirectional LSTM as the rational intent decoder. The encoder generates T hidden states by concatenating the slot output distribution $\mathbf{y}_{t}^{IS}$ and the aligned encoder hidden state ${{\mathbf{e}}_{t}}$:
\begin{equation}
{\mathbf{h}_{t}^{RI}=LSTM\left( \mathbf{h}_{t-1}^{RI},\mathbf{y}_{t-1}^{RI}\oplus \mathbf{y}_{t}^{IS}\oplus {{\mathbf{e}}_{t}} \right)}
\end{equation}
\begin{equation}
{\mathbf{y}_{t}^{RI}=\operatorname{softmax}\left( \mathbf{W}_{h}^{RI}\mathbf{h}_{t}^{RI} \right)}
\end{equation}
 where $\mathbf{h}_{t-1}^{RI}$ is the previous decoder state, $\mathbf{y}_{t}^{IS}$ is the output of intuitive slot decoder, $\mathbf{W}_{h}^{RI}$ are trainable parameters of the model, and  $\mathbf{y}_{t-1}^{RI}$ is the previously predicted slot label distribution.
\end{itemize}

\subsection{Intent2Slot Module}% 4.14修改版
As shown in Fig.~\ref{fig:network}, the Intent2Slot Module has the similar structure as the Slot2Intent Module but switches the tasks for the two decoders. In this case, we introduce a intuitive intent decoder and a rational slot decoder to explicitly incorporate intent signals to slot filling process.
\begin{itemize}
\item \textbf{Intuitive Intent Decoder.}  In practice, We  use an unidirectional LSTM as the intuitive intent decoder. The decoder state $\mathbf{h}_{t}^{II}$ of the ${{t}}$-th token is acquired by directly conditioning on the utterance representation and then utilized for producing intuitive intent features in the same way as intuitive slot decoder.
\item \textbf{Rational Slot Decoder.} Similar to the rational intent decoder, the rational slot decoder takes the concatenation of the intent output distribution $\mathbf{y}_{t}^{II}$ and the aligned encoder hidden state ${{\mathbf{e}}_{t}}$ as input units. The rational slot feature $\mathbf{h}_{t}^{RS}$ and the rational slot output distribution $\mathbf{y}_{t}^{RS}$ of the ${{t}}$-th token are obtained in the same way as rational intent decoder.
\end{itemize}
\begin{table}[b]
\caption{Dataset Statistics}
\begin{center}
\begin{tabular}{|l|c|c|}
\hline
% \textbf{Table}&\multicolumn{3}{|c|}{\textbf{Table Column Head}} \\
% \cline{2-4} 
\textbf{Dataset} & \textbf{ATIS}& \textbf{SNIPS} \\
\hline
Vocab Size& 722&11,241 \\
\hline
Average Sentence Length& 11.28&9.05 \\
\hline
\#Slots &120&72\\
\hline
\#Intents &21&7\\
\hline
\#Training Samples &4,478&13,084\\
\hline
\#Development Samples &500&700\\
\hline
\#Test Samples &893&700\\
\hline
\multicolumn{3}{l}{}
\end{tabular}
\label{dataset}
\end{center}
\end{table}
\subsection{Cooperation Mechanism}
\begin{table*}[]
\centering
\caption{Experiment results of our model and the baselines on two benchmark datasets.}%. We show the results of intent accuracy, slot F1 and overall accuracy in 
\label{tab:exp1}
\resizebox{\textwidth}{!}{%
\begin{tabular}{|l|c|c|c|c|c|c|}
\hline
\multirow{2}{*}{\textbf{Model}}            & \multicolumn{3}{c|}{\textbf{SNIPS}}           & \multicolumn{3}{c|}{\textbf{ATIS}}            \\ \cline{2-7} 
                                         & Intent (Err)  & Slot (F1)       & Overall (Acc) & Intent (Err) & Slot (F1)      & Overall (Acc) \\ \hline
Recursive NN \cite{recnn}      & 2.7          & 88.3          & -          & 4.6          & 94.0          & -          \\ 
Dilated CNN, Label-Recurrent \cite{dilatedcnn}       & 1.7          & 93.1          & -          & \textbf{1.9}          & 95.5          & -          \\ 
Attention Bi-RNN \cite{liu2016attention}   & 3.3          & 87.8            & 74.1          & 8.9          & 94.2          & 78.9          \\ 
Joint Seq2Seq \cite{hakkani2016multi} & 3.1          & 87.3          & 73.2          & 7.4          & 94.2          & 80.7          \\ 
Slot-Gated Model \cite{goo2018slot}     & 3.0            & 88.8          & 75.5            & 6.4          & 94.8          & 82.2          \\ 
Stack-Propagation \cite{qin2019stack}               & 2.0          & 94.2          & 86.9          & 3.1          & 95.9          & 86.5          \\ 
% CM-Net \cite{cmnet}               & 0.7          & 97.2          & -          & \textbf{0.9}          & 96.2          & -          \\ 
SF-ID,SF first \cite{zhang2018joint}           & 2.6          & 91.4          & 80.6          & 2.2          & 95.8           & 86.8          \\ 
SF-ID,ID first \cite{zhang2018joint}             & 2.7          & 92.2            & 80.4          & 2.9          & 95.8          & 86.9  \\ 
Graph LSTM \cite{graphlstm}             & 2.3          & 93.8            & 85.6          & 3.6          & 95.8          & 86.2 
   \\ \hline
PIN (our model)                                  & 0.9 & 94.5 & 88.0   & 2.8 & 95.9 & 87.1
\\ \hline
Joint BERT~\cite{jointbert}  & 1.4 & 97.0 & 92.8 & 2.5 & 96.1 & 88.2
\\
Graph LSTM + ELMo~\cite{graphlstm}  & 1.7 & 95.3 & 89.7 & 2.8 & 95.9 & 87.6
\\
Stack-Propagation + BERT~\cite{qin2019stack}  & 1.0 & 97.0 & 92.9 & 2.5 & 96.1 & 88.6
% CM-Net + BERT~\cite{cmnet}  & \textbf{0.7} & 97.3 & - & - & - & -
% \\
  \\ \hline
PIN(our model) + BERT                                 & \textbf{0.8} & \textbf{97.1} & \textbf{93.2}   & 2.2 & \textbf{96.3} & \textbf{88.8}
   \\ \hline
\end{tabular}%
}
\end{table*}
As described above, we have prepared the intuitive features $\mathbf{h}_{t}^{IS}$ and $\mathbf{h}_{t}^{II}$ and the rational features $\mathbf{h}_{t}^{RS}$ and $\mathbf{h}_{t}^{RI}$. It is unreasonable to treat this two different information equally when predicting the intent and slot labels. For example, in the Slot2Intent Module, the wrongly predicted slot label might mislead the rational intent decoder, causing the rational feature $\mathbf{h}_{t}^{RI}$ less reliable. In this case, the intuitive feature $\mathbf{h}_{t}^{II}$ should matter more than the rational feature $\mathbf{h}_{t}^{RI}$, which could mitigate this error propagation problem. However, if the slot labels are predicted correctly and have strong correlation with some particular intent label, $\mathbf{h}_{t}^{RI}$ is more important. Hence, we introduce a novel cooperation mechanism to make the model adaptively learn to adjust the balance:
\begin{equation}
{r_{t}^{S}=\text{ softmax}\left( \operatorname{MLP}\left( \mathbf{h}_{t}^{RS} \right) \right)}
\end{equation}
\begin{equation}
{r_{t}^{I}=\text{ softmax}\left( \operatorname{MLP}\left( \mathbf{h}_{t}^{RI} \right) \right)}
\end{equation}
\begin{equation}
{h_{t}^{S}=\mathbf{h}_{t}^{RS}\odot r_{t}^{S}+\mathbf{h}_{t}^{IS}\odot \left( 1-r_{t}^{S} \right)}
\end{equation}
\begin{equation}
{h^{I}=\operatorname{}\sum\limits_{t=1}^{T}{\mathbf{h}_{t}^{RI}\odot r_{t}^{I}+\mathbf{h}_{t}^{II}\odot \left( 1-r_{t}^{I} \right)}}
\end{equation}
where MLP represents Multilayer Perceptron, $r_{t}^{S}$ and $r_{t}^{I}$ are score vectors, $h_{t}^{S}$ and $h^{I}$ represent the fused slot and intent vectors, and $\odot $ denotes element-wise multiplication of vectors.
\subsection{Task Learning}
The final slot probability distribution of the $t$-th token and the intent probability distribution of the whole utterance are predicted by:
\begin{equation}
{\mathbf{y}_{t}^{S}=\operatorname{softmax}\left( \mathbf{W}_{h}^{S}\mathbf h_{t}^{S} \right)}
\end{equation}
\begin{equation}
{\mathbf{y}^{I}=\operatorname{softmax}\left( \mathbf{W}_{h}^{I}{\mathbf h^{I}} \right)}
\end{equation}
where $\mathbf{W}_{h}^{S}$ and $\mathbf{W}_{h}^{I}$ are the parameters to be learned.

Therefore, the cross entropy loss for SF and ID can be calculated as:
\begin{equation}
{\mathcal{L}_{slot}=\operatorname{}-\sum\limits_{j=1}^{N}\sum\limits_{t=1}^{M} l_{j}^{t,S} \log \mathbf{y}_{j}^{t,S}
}
\end{equation}
\begin{equation}
{\mathcal{L}_{intent}=\operatorname{}-\sum\limits_{j=1}^{N} l_{j}^{I} \log \mathbf{y}_{j}^{I}
}
\end{equation}
where $l_{j}^{t,S}$ and $l_{j}^{I}$ denote the ground truth label of slot and intent respectively, N is the number of training samples of each dataset, M is the number of slot labels, which depends on the length of each input utterance.

The training goal of this network is to minimize a joint loss function:
\begin{equation}
{\mathcal{L}= \lambda\mathcal{L}_{slot}+(1-\lambda)\mathcal{L}_{intent}
}
\end{equation}
where $\lambda$ is hyperparameter.
% Finally, the intent label and the slot label of the $t$-th token in the utterance are predicted by:
% % \begin{equation}
% % {\mathbf{y}_{t}^{S}=\operatorname{softmax}\left( \mathbf{W}_{h}^{S}h_{t}^{S} \right)}
% % \end{equation}
% % \begin{equation}
% % {\mathbf{y}_{t}^{I}=\operatorname{softmax}\left( \mathbf{W}_{h}^{I}h_{t}^{I} \right)}
% % \end{equation}
% \begin{equation}
% {o_{t}^{S}=\operatorname{argmax}\left(\operatorname{softmax}\left( \mathbf{W}_{h}^{S}h_{t}^{S} \right) \right)}
% \end{equation}
% \begin{equation}
% {o_{t}^{I}=\operatorname{argmax}\left(\operatorname{softmax}\left( \mathbf{W}_{h}^{I}h_{t}^{I} \right) \right)}
% \end{equation}
% where $\mathbf{W}_{h}^{S}$ and $\mathbf{W}_{h}^{I}$ are the parameters to be learned, $\mathbf{y}_{t}^{S}$ and $\mathbf{y}_{t}^{I}$ are the slot output distribution and intent output distribution of the $t$-th token in the utterance, respectively.
% Following \cite{qin2019stack}, the final sentence-level intent is generated by voting from all token-level intent results: 
% \begin{equation}
% {{{o}^{I}}=\operatorname{argmax}\sum\limits_{t=1}^{T}{\sum\limits_{j=1}^{N}{{{\alpha }_{j}}}}1\left[ o_{t}^{I}=j \right]}
% \end{equation}
% where $T$ is the length of utterance and $N$ is the number of intent labels, ${{\alpha }_{j}}$ denotes a 0-1 vector $\alpha\in \mathbb{R}^{N}$ of which the $j$-th unit is one and the others are zero, $argmax$ indicates the operation returning the indices of the maximum values in $\alpha$.
\section{Experiments and Analysis}

\subsection{Datasets}

To evaluate the effectiveness of our proposed model, we conduct experiments on two benchmark datasets: the ATIS (Airline Travel Information Systems) dataset~\cite{hemphill1990atis} and the SNIPS dataset~\cite{coucke2018snips}. Table \ref{dataset} shows the statistics of all datasets.

\begin{itemize}
\item \textbf{The ATIS dataset} contains transcribed audio recordings of people making flight reservations. It is annotated with 21 intent types and 120 slot types. Following the same data division as in \cite{goo2018slot}, we use 4,478 utterances for training set, 500 utterances for validation set and another 893 utterances for test set. 
\item \textbf{The SNIPS dataset} is a balanced real-world dataset with about 2400 utterances per each of 7 intents, which is collected from the Snips personal voice assistant. There are 72 slot types in the dataset. In this paper, we use the same SNIPS corpus setting as previous related works \cite{goo2018slot,zhang2018joint}. The training set contains 13,084 utterances, the validation set and the test set each contain 700 utterances respectively.
\end{itemize}

\subsection{Evaluation Metrics}

 Three evaluation metrics are used as our choice to measure the quality of our proposed model. The performance of intent detection is measured by intent error rate, while slot filling is evaluated with F1-score. Furthermore, the sentence accuracy is utilized to indicate the overall performance of the SLU system, which depends on the coordination between the two tasks.

\subsection{Training Details}

The model was implemented in PyTorch and trained on a single NVIDIA GeForce GTX 2080TI GPU. We preprocess the datasets following \cite{goo2018slot}. The dimension of the word embedding is 512 for SNIPS dataset and 256 for ATIS dataset. The hidden size of the encoder is set to 256. We train our model with an Adam optimizer \cite{kingma2014adam}, with a learning rate of 0.001. In order to reduce overfit, we apply L2 regularization to our model, with a decay rate of $1 \times 10^{-6}$. We set the batch  size to 16, the teacher forcing rate to 0.9, ,the dropout \cite{zaremba2014recurrent} rate to 0.4, and loss weight $\lambda$ to 0.5. During training process, early-stop strategy is performed on both datasets. 
% We also fix the random seed, so that the results are reproducible.

% \begin{table*}[t]
% \centering
% \caption{Ablation experiments on two benchmarks to investigate the impacts of various components.}
% \label{tab:exp2}
% \resizebox{\textwidth}{!}{%
% \begin{tabular}{|l|c|c|c|c|c|c|}
% \hline
% \multirow{2}{*}{\textbf{Model}}            & \multicolumn{3}{c|}{\textbf{SNIPS}}           & \multicolumn{3}{c|}{\textbf{ATIS}}            \\ \cline{2-7} 
%                                           & Slot (F1)     & Intent (Acc)  & Overall (Acc) & Slot (F1)     & Intent (Acc)  & Overall (Acc) \\  \hline
% w/o Slot2Intent Module       & \textbf{95.8}         & 96.9          & 86.5          & 95.7          & 96.9          & 86.5          \\ 
% w/o Intent2Slot Module   & 94.3          & 98.0            & 87.0          & 95.7          & 97.0          & 86.7          \\ 
% w/o Gaussian self-attention & 92.9          & 97.7          & 84.4          & 94.9          & 96.9          & 85.0          \\ 
% w/o  cooperation Fusion mechanism     & 94.3            & 98.6          & 87.4            & 95.9          & 96.6          & 87.0          \\ \hline
% Full PIN model                                  & 94.5 & \textbf{99.1} & \textbf{88.0}   & \textbf{95.9} & \textbf{97.2} & \textbf{87.1} \\ \hline
% \end{tabular}%
% }
% \end{table*}
\begin{table*}[t]
\centering
\caption{Ablation experiments on two benchmarks to investigate the impacts of various components.}
\label{tab:exp2}
\resizebox{\textwidth}{!}{%
\begin{tabular}{|l|c|c|c|c|c|c|}
\hline
\multirow{2}{*}{\textbf{Model}}            & \multicolumn{3}{c|}{\textbf{SNIPS}}           & \multicolumn{3}{c|}{\textbf{ATIS}}            \\ \cline{2-7} 
                                            & Intent (Err) & Slot (F1)     & Overall (Acc) & Intent (Err) & Slot (F1)      & Overall (Acc) \\  \hline
w/o Slot2Intent module       & 3.1         & \textbf{95.8}          & 86.5          & 3.1          & 95.7          & 86.5          \\ 
w/o Intent2Slot module   & 2.0          & 94.3            & 87.0          & 3.0          & 95.7          & 86.7          \\ 
w/o Gaussian self-attention & 2.3          & 92.9          & 84.4          & 3.1          & 94.9          & 85.0          \\ 
w/o  cooperation mechanism     & 1.4            & 94.3          & 87.4            & 3.4          & 95.9          & 87.0          \\ \hline
Full PIN model                                  & \textbf{0.9} & 94.5 & \textbf{88.0}   & \textbf{2.8} & \textbf{95.9} & \textbf{87.1} \\ \hline
\end{tabular}%
}
\end{table*}
\subsection{Baselines}

We compare our model against the following baselines:
\begin{itemize}
\item \textbf{Recursive NN}~\cite{recnn} introduced recursive neural networks to perform ID and SF together in one model, using purely lexical features and the parse tree.
\item \textbf{Dilated CNN, Label-Recurrent}~\cite{dilatedcnn} proposed a label-recurrent and dilated convolution-based model to incorporate long-distance context hierarchically for joint SF and ID. 
% incorporates long-distance context hierarchically
\item \textbf{Attention Bi-RNN}~\cite{liu2016attention} proposed an attention-based bidirectional RNN model to jointly model SF and ID. They explored various strategies to leverage explicit alignment information in the encoder-decoder models.
% leveraged the attention mechanism to allow the network to learn the relationship between slot and intent.
\item \textbf{Joint Seq2Seq}~\cite{hakkani2016multi} adopted multi-task learning method to jointly model SF and ID in a single bi-directional RNN with LSTM cells.
\item \textbf{Slot-Gated Model}~\cite{goo2018slot} introduced a slot-gated mechanism into an attention-based model to improve slot filling by the learned intent context vector. 
\item \textbf{Stack-Propagation}~\cite{qin2019stack} introduced Stack-Propagation framework to better incorporate the intent information for SF task. 
% \item \textbf{Self-Attentive Model.}~\cite{li2018self} proposed a novel self-attentive model with the intent augmented gate mechanism to utilize the semantic correlation between slot and intent.
% \item \textbf{Bi-Model.}~\cite{wang2018bi} proposed the Bi-model to consider the intent and slot filling cross-impact to each other. 
% \item \textbf{CAPSULE-NLU.}~\cite{zhang2018joint} proposed a capsule-based neural network model with a dynamic routing-by-agreement schema to accomplish slot filling and intent detection. 
\item \textbf{SF-ID Network}~\cite{haihong2019novel} proposed a SF-ID network which introduced a interrelated mechanism and iteration mechanism to establish direct connections for ID and SF to help them promote each other mutually. They designed two modes for their model: SF-First and ID-First.
\item \textbf{Graph LSTM}~\cite{graphlstm} was the ﬁrst to introduce a graph-based model to utilize the semantic correlation between slot and intent.
\item \textbf{Joint BERT}~\cite{jointbert} proposed a joint intent detection and slot ﬁlling model based on BERT.
\end{itemize}

The results of baselines are taken from \cite{recnn}, \cite{haihong2019novel} and \cite{qin2019stack} because we use the same datasets and evaluation metrics.
% For all BERT-based experiments, we just replace our utterance encoder LSTM with BERT pretrained model.

\subsection{Results}

The results across all the models are presented in Table \ref{tab:exp1} , where the proposed PIN model outperforms all the comparison methods for both datasets and almost all metrics obtain the improvement, except for the intent error rate on the ATIS.

In the SNIPS dataset, we can see that our model outperforms previous best result by 0.8\% in terms of error rate on intent detection, by 0.3\% in terms of F1-score on slot filling. It is also noteworthy that our proposed model performs better especially on sentence-level overall accuracy, where the relative improvement is around 1.1\% for SNIPS. In ATIS dataset, although we achieve top 1 performance in terms of slot F1-score and sentence-level accuracy, the relative improvement is not as obvious as the SNIPS. 
We believe the reason lies in the different complexity of the two datasets. It should be noted that, all the intents in ATIS are limited in a single domain where flight information is requested, while those in SNIPS cover multiple topics such as music and weather, adding more diversity to the SNIPS dataset. As a result, SNIPS could provide more hints about the co-occurence characteristic between intent and slot, which could be better utilized by our proposed model to boost the performances. Hence, compared with ATIS, the experimental results on SNIPS could better reflect the effectiveness of our PIN model.  Furthermore, these results can also verify our assumption that intent information and slot information can be simultaneously utilized for better promoting both two tasks.

What is more, we also investigate the effect of incorporating the pre-trained language model to our proposed method. Specifically, we substitute the PIN encoder with the BERT pretrained model and fine-tune it to further boost the performance. From the last block of Table~\ref{tab:exp1}, 
PIN + BERT outperforms all comparison approaches and establishes the new state-of-the-art performances in terms of slot F1-score and overall accuracy. We attribute this to the powerful feature extraction capabilities of BERT pre-trained model. And our model can take advantage of these enriched features to enhance the mutual guidance between ID and SF, thereby further improving performance.

\subsection{Ablation Study}

The ablation study is performed to investigate the contribution of each component in our proposed model. We remove some important components used in our model and all the variants are described as follows:
\begin{itemize}
\item W/O Slot2Intent module, where no slot information is explicitly incorporated to intent prediction process. But we still use the intuitive intent decoder from the Intent2Slot module to predict the intent of the utterance.
\item W/O Intent2Slot module, where no slot information is explicitly provided for slot filling process. Instead, we use the intuitive slot decoder from Slot2Intent module to perform SF task.
\item W/O Gaussian self-attention, where no gaussian self-attention is performed in the utterance encoder. The utterance is encoded only by BiLSTM.
\item W/O cooperation mechanism, where only the rational features are utilized for the two tasks. The other components keep the same as our model.
\end{itemize}

Table \ref{tab:exp2} presents our results on ATIS and SNIPS. We observe that all these components contribute to the superior performance of our proposed model.

If we remove the Gaussian self-attention from the holistic model, the performance on both tasks drops to some extent, and a more obvious decline can be found for the slot F1-score. We attribute it to the fact that the Gaussian self-attention is able to capture both local and contextual representation for each token, which is beneficial for slot filling task.

Besides, we also analyze the condition where either Slot2Intent Module or the Intent2Slot Module is removed. In other words, only unidirectional assistance is allowed. 
We can see from the table that the full model, which is a bidirectional one, performs better on most metrics compared with unidirectional variants. We believe the reason is that the full model does better in exploiting the ``synergy effect'' between the two tasks with its parallel structure. Specifically, our proposed model can elaborately capture the patterns of co-occurrence characteristic between slots and intents, and simultaneously provide useful guidance for the two tasks, which is conducive to the overall accuracy. 

We also examine the effect of the cooperation mechanism in Table \ref{tab:exp2}. The results show that adding the cooperation mechanism on top of the decoders can further improve SLU performance. We attribute this to the fact that by adaptively fusing intuitive features and rational features, our model can better control the proportion of contributions at each stage, which is similar to the ensemble method.

\section{Conclusion and Future Work}
In this paper, we propose a novel Parallel Interactive Network (PIN) for jointly modeling intent detection and slot filling. In our model, a Gaussian self-attentive encoder is first introduced to better capture the local context information of utterance, then two modules are introduced to model the mutual guidance between ID and SF. Finally, a cooperation mechanism is proposed to further improve the performance and robustness of our proposed PIN. Experiment results on two benchmark datasets show that the proposed PIN achieves competent performance compared with other baselines, demonstrating the effectiveness of our proposed PIN. In addition, by incorporating the pre-trained language model BERT, our method achieves the state-of-the-art among all comparison approaches.

For our future work, we will extend our model to handle cold start problem where few data samples are provided for training process.


\section*{Acknowledgment}
Special acknowledgements are given to AOTO-PKUSZ Joint Research Center for Artificial Intelligence on Scene Cognition Technology Innovation for its support.




%\bibliographystyle{IEEEtran}
\bibliographystyle{ieeetr} %ieeetr国际电气电子工程师协会期刊
\bibliography{ref} % ref就是之前建立的ref.bib文件的前缀

\end{document}